%% file: Camera_Ready.tex
\documentclass[10pt,twocolumn,letterpaper]{article}

\usepackage[pagenumbers]{cvpr} 

\usepackage{graphicx}
\usepackage{amsmath}
\usepackage{amssymb}
\usepackage{booktabs}
\usepackage{multirow}
\usepackage[accsupp]{axessibility}  

\usepackage[pagebackref,breaklinks,colorlinks]{hyperref}

\usepackage[capitalize]{cleveref}
\crefname{section}{Sec.}{Secs.}
\Crefname{section}{Section}{Sections}
\Crefname{table}{Table}{Tables}
\crefname{table}{Tab.}{Tabs.}

\def\cvprPaperID{1420} 
\def\confName{CVPR}
\def\confYear{2022}

\begin{document}

\title{CrossPoint: Self-Supervised Cross-Modal Contrastive Learning for 3D Point Cloud Understanding}

\author{Mohamed Afham$^{\dagger}$ \quad Isuru Dissanayake$^{\dagger}$ \quad Dinithi Dissanayake$^{\dagger}$ \quad Amaya Dharmasiri$^{\dagger}$ \\  Kanchana Thilakarathna$^{\ddagger}$ \quad  Ranga Rodrigo$^{\dagger}$\\
$^{\dagger}$Dept. of Electronic and Telecommunication Engineering, Univeristy of Moratuwa, Sri Lanka \\ $^{\ddagger}$The University of Sydney \\
{\tt\small afhamaflal9@gmail.com}}

\maketitle

\input{Sections/Abstract}

\input{Sections/Introduction}

\input{Sections/Related_Work_CR}

\input{Sections/Method}

\input{Sections/Experiments}

\input{Sections/Conclusion_CR}

{\small
\bibliographystyle{ieee_fullname}
\bibliography{egbib}
}

\end{document}

%% file: Sections/Abstract.tex
\begin{abstract}
    Manual annotation of large-scale point cloud dataset for varying tasks such as 3D object classification, segmentation and detection is often laborious owing to the irregular structure of point clouds. Self-supervised learning, which operates without any human labeling, is a promising approach to address this issue. We observe in the real world that humans are capable of mapping the visual concepts learnt from 2D images to understand the 3D world. Encouraged by this insight, we propose \textbf{CrossPoint}, a simple cross-modal contrastive learning approach to learn transferable 3D point cloud representations. It enables a 3D-2D correspondence of objects by maximizing agreement between point clouds and the corresponding rendered 2D image in the invariant space, while encouraging invariance to transformations in the point cloud modality. Our joint training objective combines the feature correspondences within and across modalities, thus ensembles a rich learning signal from both 3D point cloud and 2D image modalities in a self-supervised fashion. Experimental results show that our approach outperforms the previous unsupervised learning methods on a diverse range of downstream tasks including 3D object classification and segmentation. Further, the ablation studies validate the potency of our approach for a better point cloud understanding. Code and pretrained models are available at \url{https://github.com/MohamedAfham/CrossPoint}.   
\end{abstract}

%% file: Sections/Introduction.tex
\section{Introduction}
\label{sec:intro}

3D vision, which is critical in applications such as autonomous driving, mixed reality and robotics has drawn extensive attention due its ability to  understand the human world. In light of that, there have been plethora of work in 3D vision research problems such as object classification \cite{pointnet, pointnet++, dgcnn}, detection \cite{3detr} and segmentation \cite{dgcnn, pointnet++, kpconv} in the recent years with point clouds as the most popularly 3D data representation method. However, the success of deep learning crucially relies on large-scale annotated data. Even though the advancements in 3D sensing technology (e.g., LIDAR) facilitates extensive collection of 3D point cloud samples, owing to the irregular structure of point clouds, manually annotating such large-scale 3D point cloud datasets is laborious. Self-supervised learning is one of the predominant approaches to address this issue and is proven to be effective in 2D domain \cite{simclr, BYOL, AVID, dino}. 

\begin{figure}
    \centering
    \includegraphics[width = 0.99 \linewidth]{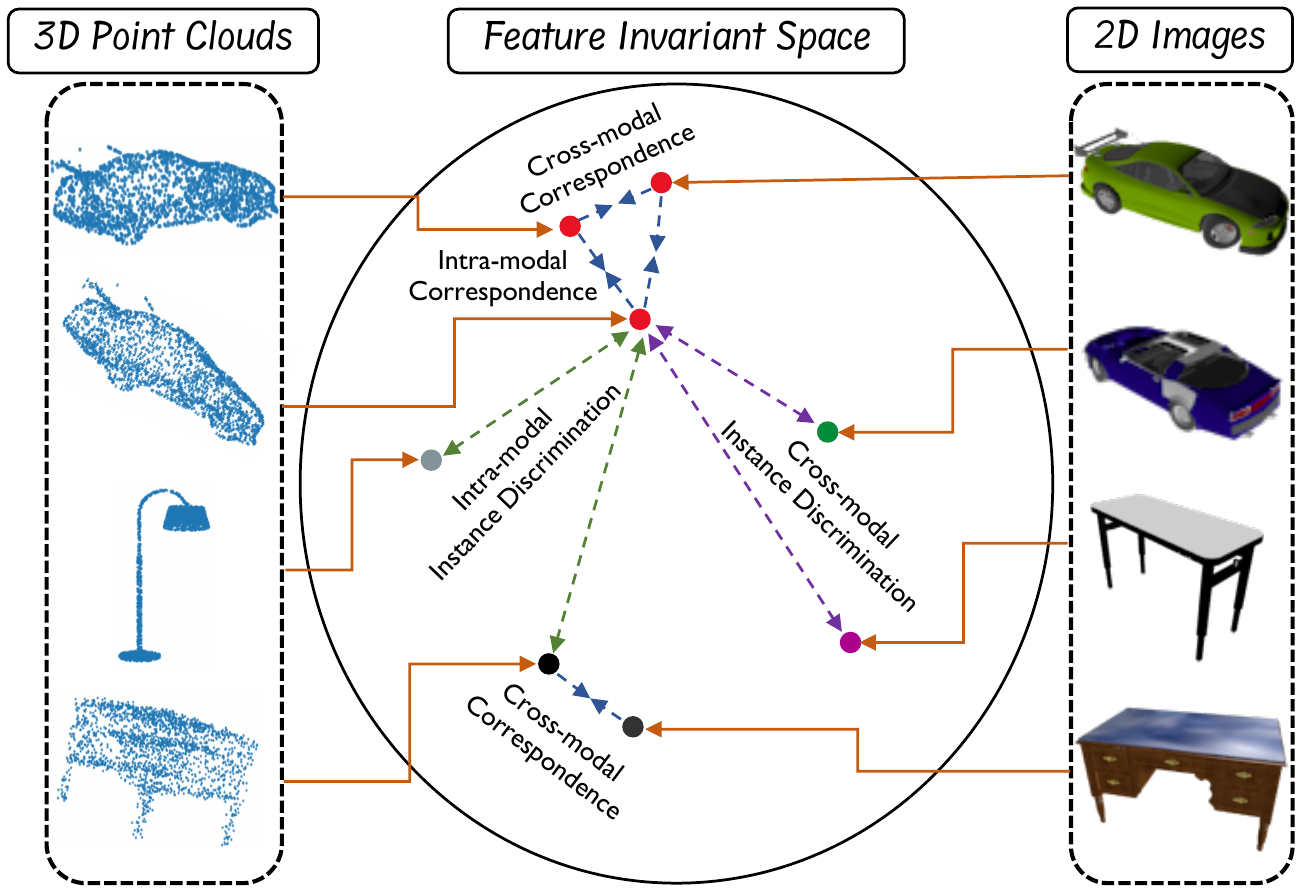}
    \caption{\textbf{An illustration of the proposed approach.} Given a 3D point cloud of an object and its rendered 2D image from a random camera view-point, \textit{CrossPoint} enforces 3D-2D correspondence while preserving the model being invariant to affine and spatial transformations via self-supervised contrastive learning. This facilitates generalizable point cloud representations which can then be utilized for 3D object classification and segmentation. Note that the 2D images shown in the right are directly rendered from the available 3D point clouds \cite{disn}.}
    \label{fig:abstract}
\end{figure}

Several works have explored self-supervised representation learning on point clouds and are mainly based on generative models \cite{3dgan, latentgan}, reconstruction \cite{jigsaw, occo} and other pretext tasks \cite{rotation}. In addition to that, with the success in exploitation of contrastive learning for image \cite{simclr, BYOL, moco, pirl} and video \cite{video1, video2, video3} understanding, recent works have investigated self-supervised contrastive learning for point cloud understanding as well \cite{strl, any_pointcloud, PointContrast, info3d}. However, the existing contrastive learning based approaches for point cloud understanding only rely on imposing invariance to augmentations of 3D point clouds. 
Learning from different modalities \textit{i.e.,} cross-modal learning has produced substantial results in self-supervised learning. Vision + language \cite{CLIP, virtex, icmlm} and video + audio \cite{AVID, listen, audiovideo} are some notable combinations for multimodal learning. Multi-modal setting has been adopted in various 2D vision tasks such as object detection \cite{mdetr}, few-shot image classification \cite{adaptive, rsfsl} and visual question and answering \cite{multimodal1, vqa}. 
Inspired by the advancements in multimodal learning, we introduce \textbf{CrossPoint}, a simple yet effective cross-modal contrastive learning approach for 3D point cloud understanding. 

The goal of our work is to capture the correspondence between 3D objects and 2D images to constructively learn transferable point cloud representations. As shown in Fig. \ref{fig:abstract}, we embed the augmented versions of point cloud and the corresponding rendered 2D image close to each other in the feature space.  In real world, humans are proficient at mapping the visual concepts learnt from 2D images to understand the 3D world. For example, a person would be able to recognize an object easily, if he/she has observed that object via an image. Cognitive scientists argue that 3D-2D correspondence is a part of visual learning process of children \cite{child1, child2}. Similarly, in real world applications such as robotics and autonomous driving, the model being aware of such 3D-2D correspondence will immensely facilitate an effective understanding of the 3D world. Our approach in particular, follows a joint objective of embedding the augmented versions of same point cloud close together in the feature space, while preserving the 3D-2D correspondence between them and the rendered 2D image of the original 3D point cloud.

The joint intra-modal and cross-modal learning objective enforces the model to attain the following desirable attributes: (a) relate the compositional patterns occurring in both point cloud and image modalities e.g., fine-grained part-level attribute of an object; (b) acquire knowledge on spatial and semantic properties of point clouds via imposing invariance to augmentations; and (c) encode the rendered 2D image feature as a centroid to augmented point cloud features thus promoting 3D-2D correspondence agnostic to transformations. Moreover, \textit{CrossPoint} does not require a memory bank for negative sampling similar to SimCLR \cite{simclr}. Formulation of rich augmentations and hard positive samples have been proved to boost the contrastive learning despite having memory banks \cite{hardpositive1, hardpositive2}. We hypothesize that the employed transformations in intra-modal setting and cross-modal correspondence provide adequate feature augmentations. In particular, the rendered 2D image feature acts as a hard positive to formulate a better representation learning. 

We validate the generalizability of our approach with multiple downstream tasks. Specifically, we perform shape classification in both synthetic \cite{modelnet} and real world \cite{scanobjectnn} object datasets. Despite being pretrained on a synthetic object dataset \cite{shapenet}, the performance of \textit{CrossPoint} in out-of-distribution data certifies the importance of the joint learning objective. In addition, the ablation studies demonstrate the component-wise contribution of both intra-modal and cross-modal objectives. We also adopt multiple widely used point cloud networks as our feature extractors, thus proving the generic nature of our approach. 

The main contributions of our approach can be summarized as follows:
\begin{itemize}
    \item We show that a simple 3D-2D correspondence of objects in the feature space using self-supervised contrastive learning facilitates an effective 3D point cloud understanding.
    \item We propose a novel end-to-end self-supervised learning objective encapsulating intra-modal and cross-modal loss functions. It encourages the 2D image feature to be embedded close to the corresponding 3D point cloud prototype, thus avoiding bias towards a particular augmentation.
    \item We extensively evaluate our proposed method across three downstream tasks namely: object classification, few-shot learning and part segmentation on a diverse range of synthetic and real-world datasets, where \textit{CrossPoint} outperforms previous unsupervised learning methods. 
    \item Additionally, we perform few-shot image classification on CIFAR-FS dataset to demonstrate that fine-tuning the pretrained image backbone from \emph{CrossPoint} outperforms the standard baseline.
    
\end{itemize}

%% file: Sections/Related_Work_CR.tex
\section{Related Work}
\label{sec:related}

\noindent \textbf{Representation Learning on Point Clouds.} Learning point cloud representation is a challenging task when compared to other modalities (e.g., images). This is because of the irregular structure and also the need for permutation invariance when processing each points. Recent line of works pioneered by PointNet \cite{pointnet} proposed methods and architectures which directly consume 3D point cloud without any preprocessing. Since then, numerous advancements have been made in point cloud based tasks such as 3D object classification \cite{pointnet++, pointcnn, kpconv, paconv, dgcnn, pointtrans, rscnn}, 3D object detection \cite{deephough, pointcnn, 3detr, h3dnet} and 3D point cloud synthesis \cite{latentgan, 3dgan}. Further, several data augmentation strategies \cite{pointaugment, pointmixup, weighted_local} have also been proposed to enhance the representation capability of the model regardless of the backbone used. However, the performance of such representation learning methods depend on the annotated point cloud data which is often hard to acquire. Sharma \etal \cite{ctree} introduced cTree, where point cloud representations can be learnt in a label efficient scenario (i.e., few-shot learning). In contrast, our approach focuses on learning transferable point cloud representations without leveraging any annotations, which can then be utilized for various downstream tasks such as classification and segmentation.

\noindent \textbf{Self - Supervised Learning on Point Clouds.}
Several approaches have been explored to perform self-supervised representation learning on point clouds. Initial line of works exploit generative modelling using generative adversarial networks \cite{3dgan, vipgan, latentgan} and auto-encoders \cite{map_vae, sonet, 3dcaps, foldingnet, gen}, which aims to reconstruct a given input point cloud with varying architectural designs. Recent line of works \cite{jigsaw, rotation, glr, mixing, multitask, occo, psg_net} introduce various pretext self-supervision tasks with the goal of learning rich semantic point attributes which eventually leads to high-level discriminative knowledge. For instance, Wang \etal \cite{occo} trains an encoder-decoder model to complete the occluded point clouds, and Poursaeed \etal \cite{rotation} defines estimation of rotation angle of the point cloud as the pretext task. However, in this work we leverage contrastive learning \cite{contrastive_loss} to learn an invariant mapping in the feature space. Inspired by the success of self-supervised contrastive learning for image understanding, numerous works \cite{PointContrast, strl, acmmm, any_pointcloud, info3d, cluster, pointdis} have analyzed such a setting for point cloud understanding. PointContrast \cite{PointContrast} performs point-level invariant mapping on two transformed view of the given point cloud. Similarly Liu \etal \cite{pointdis} also analyzes a point-level invariant mapping, by introducing point discrimination loss, which enforces the features to be consistent with points belonging to the shape surface and inconsistent with randomly sampled noisy point. STRL \cite{strl}, which is a direct extension of BYOL \cite{BYOL} to 3D point clouds, unsupervisedly learns the representations through the interactions of online and target networks. Contrary to the existing works which leverage contrastive learning, we introduce an auxiliary cross-modal contrastive objective which captures 3D-2D correspondence yielding a better representation capability.

\noindent \textbf{Cross-Modal Learning.} Learning from different modalities tend to provide rich learning signals from which semantic information of the given context can be addressed easily. Recent works \cite{CLIP, icmlm, virtex, AVID, listen, CMC} have demonstrated that pretraining in a cross-modal setting produces transferable representations which can then be deployed for various downstream tasks.
CLIP \cite{CLIP} aims to learn a multi-modal embedding space by maximizing cosine similarity between image and text modalities. Similarly, Morgado et al. \cite{AVID} combines audio and video modalities to perform a cross-modal agreement which then achieves significant gains in action recognition and sound recognition tasks. A joint learning approach with point clouds and voxels was introduced by Zhang \etal \cite{any_pointcloud}. Further, \cite{image2point} transfers the pretrained 2D image model to a point-cloud model by filter inflation. Our work is closely related to the concurrent work \cite{3d-2d}, which uses a fixed image feature extractor to perform pixel-to-point knowledge transfer. In contrast to the existing approaches, aligning with the objective of 3D understanding, CrossPoint is designed in such a way
that the 2D image feature is encouraged to be embedded close to the corresponding 3D point cloud prototype while invariance to transformations is imposed in the point cloud modality.

%% file: Sections/Method.tex
\section{Proposed Method}
\label{sec:method}

\begin{figure*}
    \centering
    \includegraphics[width=0.99 \linewidth]{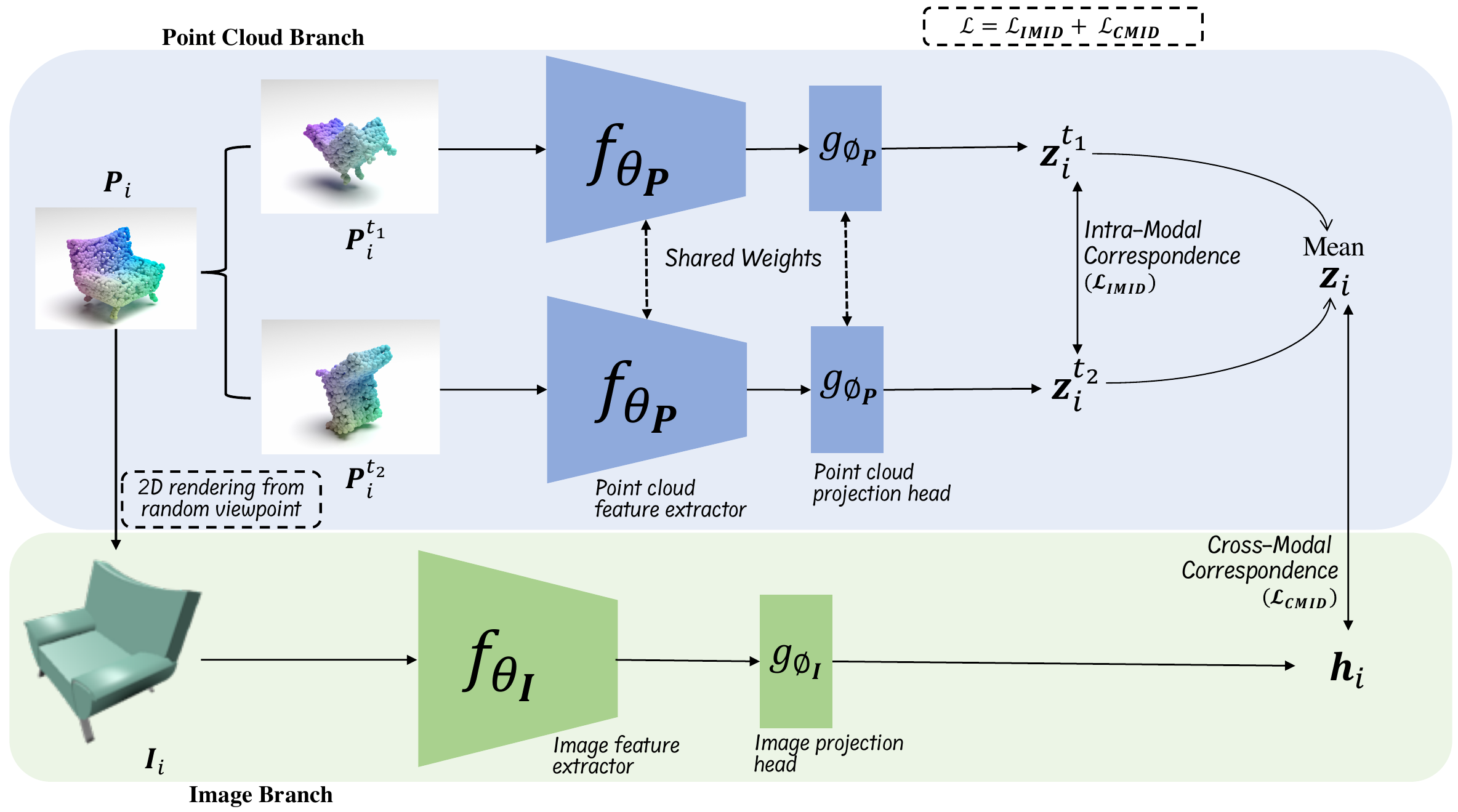}
    \caption{The overall architecture of the proposed method (CrossPoint). It comprises of two branches namely: point cloud branch which establishes an intra-modal correspondence by imposing invariance to point cloud augmentations and image branch which simply formulates a cross-modal correspondence by introducing a contrastive loss between the rendered 2D image feature and the point cloud prototype feature. CrossPoint jointly train the model combining the learning objectives of both the branches. We discard the image branch and use only the point cloud feature extractor as the backbone for the downstream tasks.}
    \label{fig:overall_architecture}
\end{figure*}

In this work, we revamp the unsupervised 3D point cloud representation learning by introducing a fusion of intra-modal and cross-modal contrastive learning objectives. 
This section begins by introducing the network architecture details of the proposed method (Sec. \ref{sec:architecture}). Then we describe the contrastive learning loss functions formulated in both intra-modal (Sec. \ref{sec:imid}) and cross-modal (Sec. \ref{sec:cmid}) settings. Finally, we present our overall training objective (Sec. \ref{sec:objective}). The overview of the proposed method is shown in Fig. \ref{fig:overall_architecture}.   

\subsection{Preliminaries} \label{sec:architecture}
Suppose we are given a dataset, $\mathcal{D} = \{\left(\textbf{P}_i, \textbf{I}_i\right)\}_{i=1}^{\vert\mathcal{D}\vert}$ with $\textbf{P}_i \in \mathbb{R}^{N \times 3}$ and $\textbf{I}_i \in \mathbb{R}^{H \times W \times 3}$ where $\textbf{I}_i$ is a rendered 2D image of the 3D point cloud $\textbf{P}_i$. Note that $\textbf{I}_i$ is obtained by capturing $\textbf{P}_i$ from a random camera view-point \cite{shapenet}. We aim to train a point cloud feature extractor $f_{\theta_\textbf{P}}\left(.\right)$ in a self-supervised manner to be effectively transferable to downstream tasks. To this end, we use an image feature extractor $f_{\theta_\textbf{I}}\left(.\right)$, multi-layer perceptron (MLP) projection heads $g_{\phi_\textbf{P}}\left(.\right)$ and $g_{\phi_\textbf{I}}\left(.\right)$ for point cloud and image respectively. 

\subsection{Intra-Modal Instance Discrimination} \label{sec:imid}

Inspired by the success of contrastive pretraining in image modality \cite{simclr, moco, pirl}, we formulate our Intra-Modal Instance Discrimination (IMID) to enforce invariance to a set of point cloud geometric transformations $\mathbf{T}$ by performing self-supervised contrastive learning. Given an input 3D point cloud $\textbf{P}_i$, we construct augmented versions $\textbf{P}_i^{t_1}$ and $\textbf{P}_i^{t_2}$ of it. We compose $t_1$ and $t_2$ by randomly combining transformations from $\mathbf{T}$ in a sequential manner. We use transformations such as rotation, scaling and translation. In addition to that we also utilize spatial transformations such as jittering, normalization and elastic distortion. Regardless of the augmentation, the corresponding transformation matrix parameters are initialized randomly.

The point cloud feature extractor $f_{\theta_\textbf{P}}$ maps both $\textbf{P}_i^{t_1}$ and $\textbf{P}_i^{t_2}$ to a feature embedding space and the resulted feature vectors are projected to an invariant space $\mathbb{R}^d$ where the contrastive loss is applied, using the projection head $g_{\phi_\textbf{P}}$. We denote the projected vectors of $\textbf{P}_i^{t_1}$ and $\textbf{P}_i^{t_2}$ as $\textbf{z}_i^{t_1}$ and $\textbf{z}_i^{t_2}$ respectively where, $\textbf{z}_i^{t} = g_{\phi_\textbf{P}}\left(f_{\theta_\textbf{P}}\left(\textbf{P}_i^{t}\right)\right)$. The goal is to maximize the similarity of $\textbf{z}_i^{t_1}$ with $\textbf{z}_i^{t_2}$ while minimizing the similarity with all the other projected vectors in the mini-batch of point clouds. We leverage NT-Xent loss proposed in SimCLR \cite{simclr} for instance discrimination at this stage. Note that our approach doesn't use any memory bank following the recent advancements in self-supervised contrastive learning \cite{BYOL, strl, dino}. We compute the loss function $l(i,{t_1},{t_2})$ for the positive pair of examples $\textbf{z}_i^{t_1}$ and $\textbf{z}_i^{t_2}$ as:

\begin{equation}
\footnotesize
\label{eq:l()}
    l(i,{t_1},{t_2}) = -\log \frac{\exp(s(\textbf{z}_i^{t_1}, \textbf{z}_i^{t_2})/\tau)}{ \sum\limits_{\substack{k=1 \\ k \neq i}}^{N} \exp(s(\textbf{z}_i^{t_1}, \textbf{z}_k^{t_1})/\tau) + \sum\limits_{\substack{k=1}}^{N} \exp(s(\textbf{z}_i^{t_1}, \textbf{z}_k^{t_2})/\tau)}
\end{equation}

where N is the mini-batch size, $\tau$ is the temperature co-efficient and $s\left(.\right)$ denotes the cosine similarity function. Our intra-modal instance discrimination loss function $\mathcal{L}_{imid}$ for a mini-batch can be described as:


\begin{equation}
\label{eq:imid}
    \mathcal{L}_{imid} = \frac{1}{2N} \sum_{i=1}^{N}[l(i,{t_1},{t_2}) + l(i,{t_2},{t_1})]
\end{equation}

\subsection{Cross-Modal Instance Discrimination} \label{sec:cmid}
In addition to the feature alignment within point cloud modality, we introduce an auxiliary contrastive objective across point cloud and image modalities to learn discriminative features, thus yielding better representation learning capability of 3D point clouds. As discussed in Sec. \ref{sec:related}, several works aim to learn transferable point cloud representations in a cross-modal setting. However, to the best of our knowledge, the joint learning objective of enforcing 3D-2D correspondence, while performing instance discrimination within point cloud modality has not been well explored. We empirically validate with the experimental results in Sec. \ref{sec:downstream} that our joint objective outperforms existing unsupervised representation methods, thus facilitating an effective representation learning of 3D point clouds.

To this end, we first embed the rendered 2D image $\textbf{I}_i$ of $\textbf{P}_i$ to a feature space using the \textit{visual backbone} $f_{\theta_\textbf{I}}$. We opt for the commonly used ResNet \cite{resnet} architecture as $f_{\theta_\textbf{I}}$.  We then project the feature vectors to the invariant space $\mathbb{R}^d$ using the image projection head $g_{\phi_\textbf{I}}$. The projected image feature is defined as $\textbf{h}_i$ where $\textbf{h}_i =  g_{\phi_\textbf{I}}\left(f_{\theta_\textbf{I}}\left(\textbf{I}_i\right)\right)$. In contrast to previous cross-modal approaches \cite{AVID, any_pointcloud}, we do not explicitly perform IMID on both the modalities (point cloud and image). Instead, we implement IMID on point cloud and leverage image modality for a better point cloud understanding. We propose a learning objective which specifically induces a bias towards 3D point cloud understanding when compared to image understanding. To this end, we compute the mean of the projected vectors $\textbf{z}_i^{t_1}$ and $\textbf{z}_i^{t_2}$ to obtain the projected prototype vector $\textbf{z}_i$ of $\textbf{P}_i$.

\begin{equation}
    \textbf{z}_i = \frac{1}{2}\left(\textbf{z}_i^{t_1} + \textbf{z}_i^{t_2} \right)
\end{equation}

In the invariance space, we aim to maximize the similarity of $\textbf{z}_i$ with $\textbf{h}_i$ since they both correspond to same objects. Our cross-modal alignment enforces the model to learn from harder positive and negative samples, thus enhances the representation capability than learning only from intra-modal alignment. We compute the loss function $l(i, \textbf{z},\textbf{h})$ for the positive pair of examples $\textbf{z}_i$ and $\textbf{h}_i$ as:

\begin{equation}
\footnotesize
\label{eq:c()}
    c(i, \textbf{z}, \textbf{h}) = -\log \frac{\exp(s(\textbf{z}_i, \textbf{h}_i)/\tau)}{ \sum\limits_{\substack{k=1 \\ k \neq i}}^{N} \exp(s(\textbf{z}_i, \textbf{z}_k)/\tau) + \sum\limits_{\substack{k=1}}^{N} \exp(s(\textbf{z}_i, \textbf{h}_k)/\tau)}
\end{equation}

where s, N, $\tau$ refers to the same parameters as in Eq. \ref{eq:l()}. The cross-modal loss function $\mathcal{L}_{cmid}$ for a mini-batch is then formulated as:

\begin{equation}
\label{eq:cmid}
    \mathcal{L}_{cmid} = \frac{1}{2N} \sum_{i=1}^{N}[c(i, \textbf{z}, \textbf{h}) + c(i, \textbf{h}, \textbf{z})]
\end{equation}


\subsection{Overall Objective} \label{sec:objective}

Finally, we obtain the resultant loss function during training as the combination of $\mathcal{L}_{imid}$ and $\mathcal{L}_{cmid}$ where $\mathcal{L}_{imid}$ imposes invariance to point cloud transformation while $\mathcal{L}_{cmid}$ injects the 3D-2D correspondence.

\begin{equation}
    \mathcal{L} = \mathcal{L}_{imid} + \mathcal{L}_{cmid}
\end{equation}

%% file: Sections/Experiments.tex
\section{Experiments}
\label{sec:exp}

\subsection{Pre-training}

\noindent \textbf{Dataset. } We use ShapeNet \cite{shapenet} as the dataset for pretraining \textit{CrossPoint}. It originally consists of more than 50,000 CAD models from 55 categories. We obtain the rendered RGB images from \cite{disn}, which has 43,783 images from 13 object categories. For a given point cloud, we randomly select a 2D image out of all the rendered images, which is captured from an arbitrary viewpoint. We use 2048 points for each point cloud while we resize the corresponding rendered RGB image to $224 \times 224.$ In addition to the augmentations applied for point cloud as described in Sec. \ref{sec:imid}, we perform random crop, color jittering and random horizontal flip for rendered images as data augmentation. \\       

\noindent \textbf{Implementation Details. }
To draw fair comparison with the existing methods, we deploy PointNet \cite{pointnet} and DGCNN \cite{dgcnn} as the point cloud feature extractors. 
We use ResNet-50 \cite{resnet} as the image feature extractor. We employ a 2-layer MLP as the projection heads which yield a 256-dimensional feature vector projected in the invariant space $\mathbb{R}^d.$ 
We use Adam \cite{adam} optimizer with weight decay $1\times10^{-4}$ and initial learning rate $1\times 10^{-3}$. Cosine annealing \cite{cosine} is employed as the learning rate scheduler and the model is trained end-to-end for 100 epochs. After pre-training we discard the image feature extractor $f_{\theta_\textbf{I}}\left(.\right)$ and projection heads $g_{\phi_\textbf{P}}\left(.\right)$ and  $g_{\phi_\textbf{I}}\left(.\right)$. All downstream tasks are performed on the pre-trained point cloud feature extractor $f_{\theta_\textbf{P}}\left(.\right)$.

\subsection{Downstream Tasks} \label{sec:downstream}
We evaluate the transferability of \textit{CrossPoint} on three widely used downstream tasks in point cloud representation learning namely: (i) 3D object classification (synthetic and real-world), (ii) Few-shot object classification (synthetic and real-world) and (iii) 3D object part segmentation.\\

\input{Tables/modelnet_cls}
\vspace{-2mm}
\noindent \textbf{(i) 3D Object classification.} We perform our classification experiments on both ModelNet40 \cite{modelnet} and ScanObjectNN \cite{scanobjectnn} to demonstrate the generalizability of our approach in both synthetic and real-world 3D shape representation learning. ModelNet40 is a synthetic dataset, where the point clouds are obtained by sampling 3D CAD models. It contains 12,331 objects (9,843 for training and 2,468 for testing) from 40 categories. ScanObjectNN \cite{scanobjectnn} is more realistic and challenging 3D point cloud classification dataset which comprises of occluded objects extractred from real-world indoor scans. It contains 2,880 objects (2304 for training and 576 for testing) from 15 categories. 

We follow the standard protocol \cite{strl, occo} to test the accuracy of our model in object classification. We freeze the pretrained point cloud feature extractor and fit a simple linear SVM classifier on the train split of the classification datasets. We randomly sample 1024 points from each object for both training and testing the classification results. Our CrossPoint also delivers a consistent performance in different backbones. We perform experiments in both PointNet \cite{pointnet} and DGCNN \cite{dgcnn} where PointNet is an MLP based feature extractor while DGCNN is built on graph convolutional networks. Table. \ref{table:modelnet_cls} reports the linear classification results on ModelNet40. It is clear that CrossPoint outperforms previous state-of-the-art unsupervised methods in both of the feature extractors, thus establishing a new benchmark in self-supervised learning for point clouds. In particular, our model outperforms DepthContrast \cite{any_pointcloud}, which also employ a cross-modal setting for point cloud representation learning, by a significant margin of 5.8\%. While our approach surpasses prior works which utilizes self-supervised contrastive learning \cite{strl, multitask, acmmm} by considerable margins, some other methods \cite{pointdis, PointContrast, glr} cannot be compared in a fair manner for varying reasons such as different pretraining mechanisms and discrepancy in feature extractors.

\input{Tables/scanobjectnn_cls}

Table. \ref{table:scanobjectnn_cls} demonstrates the linear evaluation results on ScanObjectNN. Significant accuracy gains of 1.3\% in PointNet backbone and 3.4\% in DGCNN backbone, over previous state-of-the-art unsupervised methods show that our proposed joint learning approach generalizes to out-of-distribution data as well. 

\input{Tables/fsl}
\vspace{-4mm}
\paragraph{\textbf{(ii) Few-shot object classification.}} Few-shot learning (FSL) aims to train a model that generalizes with limited data. We conduct experiments on conventional few-shot task (N-way K-shot learning), where the model is evaluated on N classes, and each class contains N samples. Several works in 2D domain \cite{rfs, exploring} have shown the effectiveness of training a simple linear classifier on top of representation trained in a self-supervised manner. Similar to standard 3D object classification, we use ModelNet40 and ScanObjectNN datasets to carry out FSL experiments. While there is no standard split for FSL in both of the datasets, for a fair comparison with previous methods \cite{occo, ctree}, we randomly sample 10 few-shot tasks and report the mean and standard deviation. Table. \ref{tab:fsl} shows the FSL results on ModelNet40, where CrossPoint outperforms prior works in all the FSL settings in both PointNet and DGCNN backbones. It is noticeable that our method with DGCNN backbone performs poorly in some of the FSL settings compared to that with PointNet backbone. Similar pattern is also observed in previous methods \cite{occo, ctree} as well. We attribute to the fact that complex backbones might degrade the few-shot learning performance which has been observed consistently in the FSL literature in images \cite{rfs}.

We report the FSL results on ScanObjectNN dataset in Table. \ref{tab:scan_fsl}. CrossPoint produces significant accuracy gains in most of the settings in both PointNet and DGCNN feature extractors, proving the capability of generalizing with a limited data even in an out-of-distribution setting.

\input{Tables/scanobjectnn_fsl}
\vspace{-5mm}
\paragraph{\textbf{(iii) 3D Object part segmentation.}} We perform object part segmentation in the widely used ShapeNetPart dataset \cite{shapenetpart}. It contains 16881 3D objects from 16 categories, annotated with 50 parts in total. We initially pretrain the backbone proposed in DGCNN \cite{dgcnn} for part segmentation using our approach in ShapeNet dataset and fine tune in an end-to-end manner in the train split of ShapeNetPart dataset. We report the mean IoU (Intersection-over-Union) metric, calculated by averaging IoUs for each part in an object before averaging the obtained values for each object class in Table. \ref{tab:partseg}. Part segmentation using the backbone pretrained via CrossPoint outperforms the randomly initialised DGCNN backbone by 0.4\%. This shows that CrossPoint provides a better weight initialization to the feature extractors. Accuracy gains over the previous self-supervised learning frameworks indicates that CrossPoint, by imposing intra-modal and cross-modal correspondence in a joint manner, tends to capture fine-grained part-level attributes which is crucial in part segmentation.

\input{Tables/partseg}

\subsection{Ablations and Analysis} \label{sec:analysis}

\begin{figure}[h]
    \centering
    \includegraphics[width = 0.99 \linewidth]{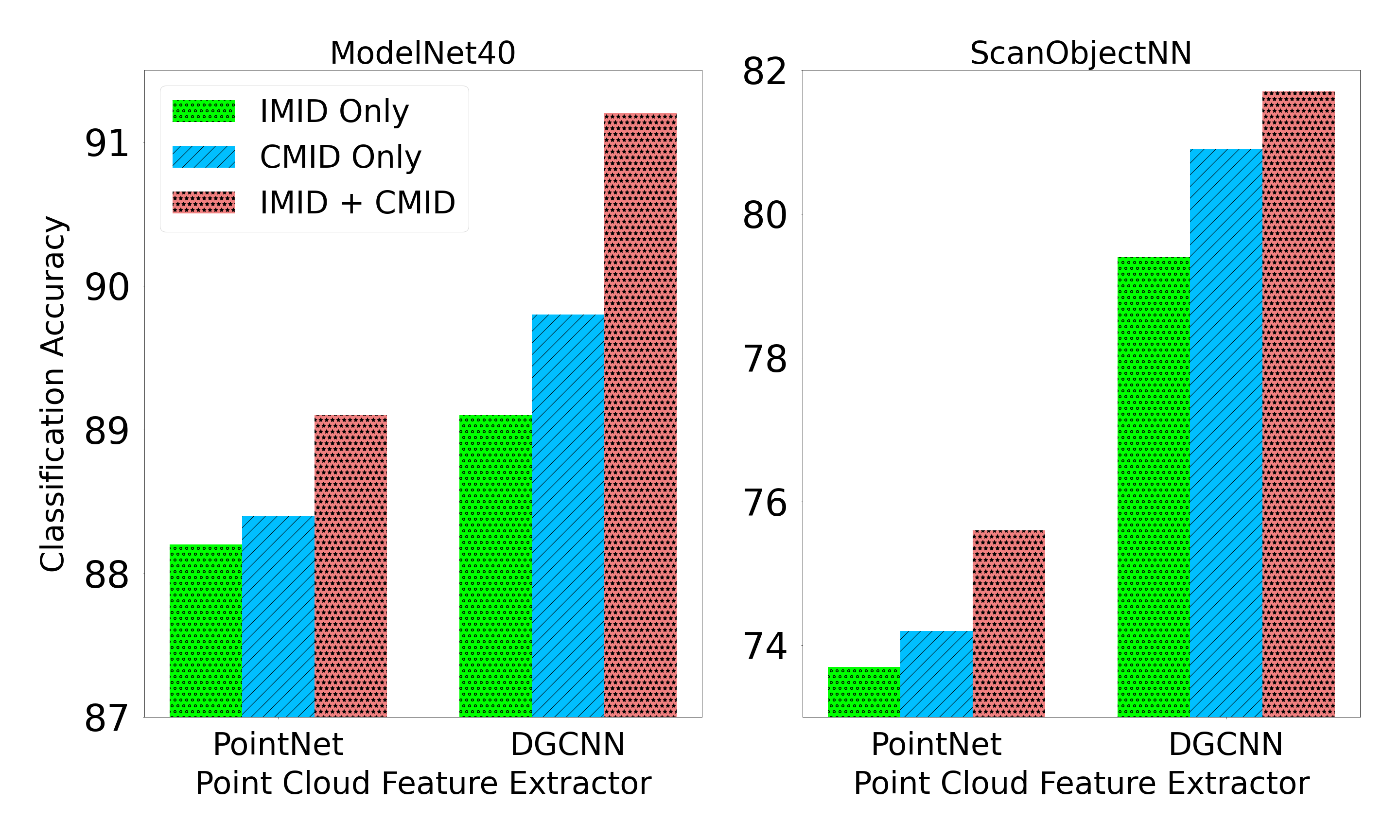}
    \caption{Impact of the joint learning objective when compared to individual intra-modal and cross-modal objectives. Classification results with Linear SVM on the pretrained embedding on ModelNet40 (left) and ScanObjectNN (right) dataset.}
    \label{fig:loss_ablation}
\end{figure}

\noindent \textbf{Impact of joint learning objective.} As described in Sec. \ref{sec:method}, our approach aims to train the model with a joint learning objective. We hypothesize that, addressing both intra-modal and cross-modal correspondence in a joint manner contribute to a better representation learning than the individual learning objectives. Intra-modal correspondence encourages the model to capture the fine-grained part semantics via imposing invariance to transformations and cross-modal correspondence establishes hard positive feature samples for contrastive learning to make the learning even more challenging, thus yielding better results. We empirically test this hypothesis by training the model in all possible settings and evaluating a linear SVM classifier in both ModelNet40 and ScanObjectNN datasets. Figure. \ref{fig:loss_ablation} graphically illustrates that in all the learning settings, the proposed joint learning paradigm performs better than the individual objectives. In particular, the combination of intra-modal and cross-modal learning objectives obtains accuracy gains of 1.2\% and 0.7\% over the second best approach in ModelNet40 and ScanObjectNN respectively with the DGCNN feature extractor.

\begin{figure*}
    \centering
    \includegraphics[width = 0.99 \linewidth]{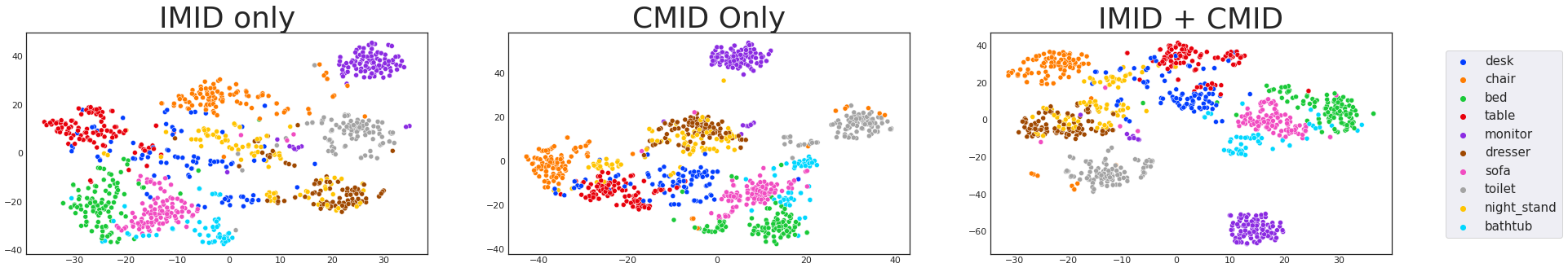}
    \caption{t-SNE visualization of features on the test split of \textbf{ModelNet10} dataset after training the DGCNN backbone in a self-supervised manner. The proposed joint learning approach provide better discrimination of classes (e.g., \textit{desk, table}) when compared to models learnt with individual objectives.}
    \label{fig:tsne}
\end{figure*}

We specifically observe that linear evaluation with cross-modal learning objective marginally outperforms that with intra-modal learning objective in the classification accuracy metrics. We believe that the cross-modal learning objective facilitates a part semantic understanding, by embedding the image feature in the close proximity to the features of both the augmented point clouds by leveraging the point cloud prototype feature. Figure. \ref{fig:tsne} visualizes the t-SNE plot of the features obtained from the test split of ModelNet10 dataset. It is visible that both CMID and IMID settings provide a good discrimination to classes even without explicitly trained with labeled data. However, the class boundaries of some classes (e.g., desk, table) are not precise and compact. Joint learning objective is able to create a better discrimination boundaries in such classes. 

\noindent \textbf{Number of corresponding 2D images.} We investigate the contribution of the image branch by varying the number of rendered 2D images (n). We select the rendered 2D image captured from different random directions. In case of more than one rendered 2D image, we compute the mean of all the projected features of rendered images to perform the cross-modal instance discrimination (CMID). Table. \ref{tab:multiple_images} reports the linear SVM classification results on ModelNet40 dataset. Our approach, even with a single rendered 2D image captures the cross-modal correspondence to yield better linear classification results. It is clear that, when using more than 2 rendered images, the information gathered from 2D image modality might have become redundant, hence the drop in accuracy.

\begin{table}[h]
    \centering
    \caption{\textbf{Linear classification results on ModelNet40 with varying number of rendered 2D images (n).} CrossPoint with single corresponding image performs better than or equal to multiple rendered images. We choose n=1 for all the experiments.}
    \label{tab:multiple_images}
    \begin{tabular}{cccccc}
    \toprule
    \begin{tabular}{l}No. of rendered\\2D images (n)\end{tabular} & \textbf{1} & \textbf{2} & \textbf{3} & \textbf{4} & \textbf{5}\\ \hline
    Linear Accuracy & \textbf{91.2} & 91.2 & 90.9 & 91.0 & 90.5 \\ 
    \bottomrule
    \end{tabular}
\end{table}

\noindent \textbf{Few-shot image classification on CIFAR-FS.} Even though we discard the image feature extractor during point cloud downstream tasks, we perform a simple few-shot image classification to investigate the image understanding capability of it. We use CIFAR-FS \cite{cifar}, which is a widely used dataset for few-shot image classification, that contains 100 categories with 64, 16, and 20 train, validation, and test splits. Table. \ref{tab:cifar} reports the results in comparison with the standard baseline, RFS \cite{rfs} in 5-way 1-shot and 5-way 5-shot settings. It is to be noticed that CrossPoint, without any supervised fine-tuning, fails to generalize well in the few-shot image classification setting. We believe that this is because, there is a considerable discrepancy between the rendered 2D images from point clouds and the images in CIFAR-FS which are real world images. Hence, CrossPoint fails to generalize to such an out-of-distribution data which is a limitation of our work. However, initializing the backbone with the unsupervisedly trained image feature extractor in CrossPoint and finetuning using the method proposed in RFS outperforms the baseline results by significant margins in both the few-shot settings. 

\begin{table}[h]
\centering
\small
\setlength{\tabcolsep}{3pt}
\caption{\textbf{Few-shot image classification results on CIFAR-FS.} Fine-tuning CrossPoint with RFS improves the performance.}
\label{tab:cifar}
\begin{tabular}{lcccc}
\toprule 
Method  & Backbone & 5-way 1-shot & 5-way 5-shot \\ \bottomrule
CrossPoint & ResNet-50 & 24.12$\pm$0.48 & 28.18$\pm$0.54\\
RFS \cite{rfs} & ResNet-50 & 60.20$\pm$0.87 & 76.79$\pm$0.71\\
CrossPoint + RFS & ResNet-50 & \textbf{64.45$\pm$0.86} & \textbf{80.14$\pm$0.65}\\
\bottomrule
\end{tabular}
\end{table}

%% file: Tables/modelnet_cls.tex
\begin{table}[t]
\centering \small
\caption{\textbf{Comparison of ModelNet40 linear classification results with previous self-supervised methods.} A linear classifier is fit onto the training split of ModelNet40 using the pretrained model and overall accuracy for classification in test split is reported. Our method CrossPoint, surpasses existing works in both PointNet and DGCNN backbones.}
\label{table:modelnet_cls}
\begin{tabular}{l c c} 
 \toprule 
 Method & ModelNet40 \\
 \midrule
 3D-GAN \cite{3dgan}& 83.3 \\
 Latent-GAN \cite{latentgan}  & 85.7 \\
 SO-Net \cite{sonet} & 87.3 \\
 FoldingNet \cite{foldingnet} & 88.4 \\
 MRTNet \cite{mrtnet} & 86.4 \\
 3D-PointCapsNet \cite{3dcaps} & 88.9 \\
 DepthContrast \cite{any_pointcloud} & 85.4 \\
 ClusterNet \cite{cluster} & 86.8 \\
 VIP-GAN \cite{vipgan} & 90.2 \\
 \hline
 PointNet + Jigsaw \cite{jigsaw} & 87.3 \\
 PointNet + STRL \cite{strl} & 88.3 \\
 PointNet + Rotation \cite{rotation} & 88.6 \\
 PointNet + OcCo \cite{occo} & 88.7 \\
 \textbf{PointNet + CrossPoint (Ours)} & \textbf{89.1} \\
 \hline
 DGCNN + Multi-Task \cite{multitask} & 89.1 \\
 DGCNN + Self-Contrast \cite{acmmm} & 89.6 \\
 DGCNN + Jigsaw \cite{jigsaw} & 90.6 \\
 DGCNN + STRL \cite{strl} &  90.9 \\
 DGCNN + Rotation \cite{rotation} & 90.8 \\
 DGCNN + OcCo \cite{occo} & 89.2 \\
 \textbf{DGCNN + CrossPoint (Ours)} & \textbf{91.2} \\
 \bottomrule
\end{tabular}
\end{table}

%% file: Tables/scanobjectnn_cls.tex
\begin{table}[h]
\centering \small
\caption{\textbf{Comparison of ScanObjectNN linear classification results with previous self-supervised methods.} CrossPoint shows consistent improvements over prior works in both PointNet and DGCNN backbones. This illustrates the effectiveness of our approach in real-world setting.}
\label{table:scanobjectnn_cls}
\begin{tabular}{l|cc}
    \bottomrule
\multicolumn{1}{c|}{\multirow{2}{*}{Method}}
        & \multicolumn{2}{c}{Backbone} \\
    \cline{2-3}
        & PointNet  & DGCNN\\
    \toprule \bottomrule
    Jigsaw \cite{jigsaw} & 55.2 & 59.5 \\
    OcCo \cite{occo} & 69.5 & 78.3 \\
    STRL \cite{strl} & 74.2 & 77.9 \\
    \textbf{CrossPoint (Ours)} & \textbf{75.6} & \textbf{81.7}\\
    \toprule
\end{tabular}
\end{table}

%% file: Tables/fsl.tex
\begin{table}[t!]
\centering
\footnotesize
\setlength{\tabcolsep}{3pt}
\caption{\textbf{Few-shot object classification results on ModelNet40.} We report mean and standard error over 10 runs. Top results of each backbone is colored in red and blue. Proposed CrossPoint improves the few-shot accuracy in all the reported settings. Table is an extended version from \cite{occo}}
\label{tab:fsl}
\begin{tabular}{l|cccc}
\bottomrule \multicolumn{1}{c|}{\multirow{2}{*}{Method}}  & 
\multicolumn{2}{c}{5-way} & \multicolumn{2}{c}{10-way} \\\cline{2-5}
& 10-shot & 20-shot & 10-shot & 20-shot \\ \toprule \bottomrule
3D-GAN \cite{3dgan} & 55.8$\pm$3.4 & 65.8$\pm$3.1 & 40.3$\pm$2.1 & 48.4$\pm$1.8 \\
FoldingNet \cite{foldingnet}  & 33.4$\pm$4.1 & 35.8$\pm$5.8 & 18.6$\pm$1.8 & 15.4$\pm$2.2 \\
Latent-GAN \cite{latentgan}  & 41.6$\pm$5.3 & 46.2$\pm$6.2 & 32.9$\pm$2.9 & 25.5$\pm$3.2 \\
3D-PointCapsNet \cite{3dcaps} & 42.3$\pm$5.5 & 53.0$\pm$5.9 & 38.0$\pm$4.5 & 27.2$\pm$4.7 \\
PointNet++ \cite{pointnet++}  & 38.5$\pm$4.4 & 42.4$\pm$4.5 & 23.1$\pm$2.2 & 18.8$\pm$1.7 \\
PointCNN \cite{pointcnn}& 65.4$\pm$2.8 & 68.6$\pm$2.2 & 46.6$\pm$1.5 & 50.0$\pm$2.3 \\
RSCNN \cite{rscnn} & {65.4$\pm$8.9} & {68.6$\pm$7.0} & {46.6$\pm$4.8} & {50.0$\pm$7.2} \\ \hline
PointNet + Rand & 52.0$\pm$3.8 & 57.8$\pm$4.9 & 46.6$\pm$4.3 & 35.2$\pm$4.8 \\
PointNet + Jigsaw \cite{jigsaw} & 66.5$\pm$2.5 & 69.2$\pm$2.4 & 56.9$\pm$2.5 & 66.5$\pm$1.4\\
PointNet + cTree \cite{ctree}& 63.2$\pm$3.4 & 68.9$\pm$3.0 & 49.2$\pm$1.9 & 50.1$\pm$1.6 \\
PointNet + OcCo \cite{occo}& 89.7$\pm$1.9 & 92.4$\pm$1.6 & 83.9$\pm$1.8 & 89.7$\pm$1.5 \\
\textbf{PointNet + CrossPoint} & \textcolor{red}{90.9$\pm$4.8} &\textcolor{red}{93.5$\pm$4.4} & \textcolor{red}{84.6$\pm$4.7} & \textcolor{red}{90.2$\pm$2.2} \\\hline
DGCNN + Rand & 31.6$\pm$2.8 & 40.8$\pm$4.6 & 19.9$\pm$2.1 & 16.9$\pm$1.5\\
DGCNN + Jigsaw \cite{jigsaw}& 34.3$\pm$1.3 & 42.2$\pm$3.5 & 26.0$\pm$2.4 & 29.9$\pm$2.6\\
DGCNN + cTree \cite{ctree} & 60.0$\pm$2.8 & 65.7$\pm$2.6 & 48.5$\pm$1.8 & 53.0$\pm$1.3\\
DGCNN + OcCo \cite{occo} & 90.6$\pm$2.8 & 92.5$\pm$1.9 & 82.9$\pm$1.3 & 86.5$\pm$2.2 \\
\textbf{DGCNN + CrossPoint} & \textcolor{blue}{92.5$\pm$3.0} &\textcolor{blue}{94.9$\pm$2.1} & \textcolor{blue}{83.6$\pm$5.3} & \textcolor{blue}{87.9$\pm$4.2}\\
\toprule \bottomrule
\end{tabular}
\end{table}

%% file: Tables/scanobjectnn_fsl.tex
\begin{table}[h]
\centering
\footnotesize
\setlength{\tabcolsep}{3pt}
\caption{\textbf{Few-shot object classification results on ScanObjectNN.} We report mean and standard error over 10 runs. Top results of each backbone is colored in red and blue. Proposed CrossPoint improves the few-shot accuracy in all the reported settings. Table is an extended version from \cite{occo}}
\label{tab:scan_fsl}
\begin{tabular}{l|cccc}
\bottomrule \multicolumn{1}{c|}{\multirow{2}{*}{Method}}  & 
\multicolumn{2}{c}{5-way} & \multicolumn{2}{c}{10-way} \\\cline{2-5}
& 10-shot & 20-shot & 10-shot & 20-shot \\ \toprule \bottomrule
PointNet + Rand & 57.6$\pm$2.5 & 61.4$\pm$2.4 & 41.3$\pm$1.3 & 43.8$\pm$1.9 \\
PointNet + Jigsaw \cite{jigsaw} & 58.6$\pm$1.9 & 67.6$\pm$2.1 & 53.6$\pm$1.7 & 48.1$\pm$1.9\\
PointNet + cTree \cite{ctree}& 59.6$\pm$2.3 & 61.4$\pm$1.4 & 53.0$\pm$1.9 & 50.9$\pm$2.1 \\
PointNet + OcCo \cite{occo}& \textcolor{red}{70.4$\pm$3.3} & 72.2$\pm$3.0 & 54.8$\pm$1.3 & 61.8$\pm$1.2 \\
\textbf{PointNet + CrossPoint} & 68.2$\pm$1.8 & \textcolor{red}{73.3$\pm$2.9} & \textcolor{red}{58.7$\pm$1.8} & \textcolor{red}{64.6$\pm$1.2} \\\hline
DGCNN + Rand & 62.0$\pm$5.6 & 67.8$\pm$5.1 & 37.8$\pm$4.3 & 41.8$\pm$2.4\\
DGCNN + Jigsaw \cite{jigsaw}& 65.2$\pm$3.8 & 72.2$\pm$2.7 & 45.6$\pm$3.1 & 48.2$\pm$2.8\\
DGCNN + cTree \cite{ctree} & 68.4$\pm$3.4 & 71.6$\pm$2.9 & 42.4$\pm$2.7 & 43.0$\pm$3.0\\
DGCNN + OcCo \cite{occo} & 72.4$\pm$1.4 & 77.2$\pm$1.4 & 57.0$\pm$1.3 & 61.6$\pm$1.2\\
\textbf{DGCNN + CrossPoint} & \textcolor{blue}{74.8$\pm$1.5} &\textcolor{blue}{79.0$\pm$1.2} & \textcolor{blue}{62.9$\pm$1.7} & \textcolor{blue}{73.9$\pm$2.2}\\
\toprule \bottomrule
\end{tabular}
\end{table}

%% file: Tables/partseg.tex
\begin{table}[h]
    \centering
    \caption{\textbf{Part segmentation results on ShapeNetPart dataset.} We report the mean IoU across all the object classes. \textit{Supervised} indicates the models trained with randomly initialising feature backbones, while \textit{Self-Supervised} models are the ones initialised with pretrained feature extractors.}
    \label{tab:partseg}
    \begin{tabular}{llc}
    \toprule
        Category & Method & Mean IoU \\ \bottomrule
        \multirow{3}{*}{\textit{Supervised}} & PointNet \cite{pointnet} & 83.7 \\
        & PointNet++ \cite{pointnet++} & 85.1 \\
        & DGCNN \cite{dgcnn} & 85.1 \\
        \hline
        \multirow{6}{*}{\textit{Self-Supervised}} & Self-Contrast \cite{acmmm} & 82.3 \\
        & Jigsaw \cite{jigsaw} & 85.3 \\
        & OcCo \cite{occo} & 85.0 \\
        & PointContrast \cite{PointContrast} & 85.1 \\
        & Liu \etal \cite{pointdis} & 85.3 \\
        & \textbf{CrossPoint (Ours)} & \textbf{85.5} \\
        \bottomrule
    \end{tabular}
\end{table}

%% file: Sections/Conclusion_CR.tex


\section{Conclusion}
\label{sec:conclusion}

In this paper, we propose CrossPoint, a simple self-supervised learning framework for 3D point cloud representation learning. Even though our approach is trained on synthetic 3D object dataset, experimental results in downstream tasks such as 3D object classification and 3D object part segmentation in both synthetic and real-world datasets demonstrate the efficacy of our approach in learning transferable representations. Our ablations empirically validate our claim that joint learning of imposing intra-modal and cross-modal correspondences leads to more generic and transferable point cloud features. Additional few-shot image classification experiment provides a powerful insight for cross-modal understanding which can be explored in future researches. 


\section*{Acknowledgments}
The authors would like to thank Salman Khan (MBZUAI, UAE) and Sadeep Jayasumana (Google Research, NY) for their valuable comments and suggestions on the manuscript.